\documentclass[sigconf]{acmart}
\usepackage{subfig}
\usepackage{verbatim}
\usepackage{bm}
\usepackage{array}

\AtBeginDocument{%
  }

\setcopyright{acmcopyright}
\copyrightyear{2023}
\acmYear{2023}
\acmDOI{XXXXXXX.XXXXXXX}

\acmConference[XXX]{XXX}{XXX}{XXX}
\acmPrice{15.00}
\acmISBN{978-1-4503-XXXX-X/18/06}

\begin{document}

\title{Teach LLMs to Personalize -- An Approach inspired by Writing Education}


\author{Cheng Li}
\affiliation{%
  \institution{Google}
  \country{USA}
}
\email{chgli@google.com}

\author{Mingyang Zhang}
\affiliation{%
  \institution{Google}
  \country{USA}
}
\email{mingyang@google.com}

\author{Qiaozhu Mei}
\authornote{Work done as a visiting researcher at Google.}
\affiliation{%
  \institution{University of Michigan}
  \country{USA}
}
\email{qmei@umich.edu}

\author{Yaqing Wang}
\affiliation{%
  \institution{Google}
  \country{USA}
}
\email{yaqingwang@google.com}

\author{Spurthi Amba Hombaiah}
\affiliation{%
  \institution{Google}
  \country{USA}
}
\email{spurthiah@google.com}

\author{Yi Liang}
\affiliation{%
  \institution{Google}
  \country{USA}
}
\email{yiliang@google.com}

\author{Michael Bendersky}
\affiliation{%
  \institution{Google}
  \country{USA}
}
\email{bemike@google.com}


\renewcommand{\shortauthors}{Li et al.}

\begin{abstract}
Personalized text generation is an emerging research area that has attracted much attention in recent years. Most studies in this direction focus on a particular domain by designing bespoke features or models. In this work, we propose a general approach for personalized text generation using large language models (LLMs). Inspired by the practice of writing education, we develop a multistage and multitask framework to teach LLMs for personalized generation.
In writing instruction, the task of writing from sources is often decomposed into multiple steps that involve finding, evaluating, summarizing, synthesizing, and integrating information. Analogously, our approach to personalized text generation consists of multiple stages: retrieval, ranking, summarization, synthesis, and generation. In addition, we introduce a multitask setting that helps the model improve its generation ability further, which is inspired by the observation in education that a student's reading proficiency and writing ability are often correlated.
We evaluate our approach on three public datasets, each of which covers a different and representative domain. Our results show significant improvements over a variety of baselines.

\end{abstract}




\maketitle

\section{Introduction}
As artificial intelligence (AI) based systems are increasingly used to assist content creation, there has been a tremendous amount of interest in personalized text generation. Producing a customized response that takes into account auxiliary context, such as documents previously written by the user, is crucial for the development of generative systems that support specific audiences, creation contexts, and information needs. Example applications include AI-assisted writing of various types of content from tweets and news stories to scientific articles and fictions, corporate and personal communications (emails, chats, forums), and transformations of a given piece of written content into other styles, e.g., summarization, or conversely, elaboration.

Researchers have investigated the generation of personalized text on various domains, including but not limited to reviews~\cite{li2019towards, li2020knowledge}, dialogue agents~\cite{wu2021personalized, zhang2018personalizing, mazare2018training} and social networks~\cite{gaoresearch}. Previous work mostly relies on domain-specific features or knowledge and proposes models that address a particular task. How to design a general approach that works for all scenarios is a less studied area.

On the other hand, along with the ascendance of generative AI, through chatbots like ChatGPT\footnote{\url{https://chat.openai.com}} and Bard\footnote{\url{https://bard.google.com}} in particular, large language models (LLMs) are playing an increasingly prominent role in many text generation tasks. However, few studies have explored how to equip LLMs with the ability to personalize.

In this work, we propose a general approach for personalized text generation using large language models. Our work is inspired by the widely-used practice in writing education, which decomposes the task of writing from sources by a procedure of finding, evaluating, summarizing, synthesizing, and integrating information~\cite{shanahan2015common, cooney2018integrating}. Analogously, we adopt a multistage multitask framework to teach LLMs for personalized text generation, with similar stages being retrieval, ranking, summarization, synthesis, and generation. Specifically, given the immediate context, such as the title and the starting sentence of a document a user is writing, we formulate a query and \textit{retrieve} relevant information from an auxiliary repository of personal contexts, such as documents the user has authored in the past. We then \textit{rank} the retrieved results based on their relevance and importance, followed by \textit{summarizing} the ranked results. We also \textit{synthesize} the retrieved information into key elements, and finally feed the retrieved results, summary and synthesized information into the large language model for generating the new document.

In language education, it is often observed that the proficiency of one's writing skills is highly correlated with that of their reading skills~\cite{crowhurst1990reading}. Furthermore, studies show that author recognition tasks can be used to measure the amount and level of reading by an individual~\cite{mccarron2021author}, which correlates with their reading proficiency. Inspired by these two observations, we create a multitask setting that aims to improve the reading ability of the large language model, where we introduce an auxiliary task charging the model to attribute the authorship of a given text. We anticipate that this task will help the model better understand (i.e., read) the given text and in turn generate (i.e., write) better and more personalized content.

We evaluate the proposed models on three public datasets, which cover personal email communications, social media discussions, and product reviews. By employing our multistage and multitask framework, we demonstrate significant improvements over a variety of baselines on all three datasets.

\section{Related Work}
\label{sec:related}
We present a literature review on personalized text generation and two related tasks, controlled text generation and text style transfer.

\textit{Personalized text generation.}
Some studies focus on improving personalized generation for a particular domain by utilizing domain-specific features or knowledge. Li and Tuzhilin~\cite{li2019towards} design a model based on self-attentive recursive autoencoders to generate personalized user reviews given product description, sentiment labels, and historical reviews of the user. A knowledge enhanced personalized review generation model based on a capsule graph neural network (CapsGNN) is proposed in~\cite{li2020knowledge} to utilize product attributes. Gao et al.~\cite{gaoresearch} focus on personalized social text generation, where personalized features are fed to the encoder to guide the generation of the decoder. There are extensive studies on personalization for dialogue agents~\cite{wu2021personalized, zhang2018personalizing, mazare2018training}. Due to limited real conversational data, researchers have explored constructing data by asking crowd-workers to write dialogues for specific personas~\cite{zhang2018personalizing} and by extracting user attributes and utterances from Reddit~\cite{wu2021personalized, mazare2018training} and Weibo~\cite{zhong2022less, qian2021pchatbot}.

There are investigations on using predefined attributes and topics for personalization. A personalized sentence generation method is proposed~\cite{yuan2020personalized} based on generative adversarial networks (GANs). Frequently used function words and content words are used as input and as sentence structure constraints for model training.

A less explored area is how to utilize large language models for personalized generation across different domains without relying on domain-specific or user-defined features. LaMP~\cite{salemi2023lamp} is the work closest to ours. It provides a benchmark for training and evaluating personalized language models on three classification and four text generation tasks. They deploy an approach that retrieves text from user profiles. The generation tasks provided in LaMP are at the sentence-level. We instead consider generating longer text of passage-length, which is more challenging. Method-wise, the retrieval based approach in LaMP can be viewed as an instantiation of a single component of the multi-stage framework we proposed. Skopyk et al.~\cite{skopykpersonalizing} propose to train transformer layer adapters to achieve the effect of personalization. The paper only proposes the method without including any experimental analysis.

\textit{Controlled text generation.}
Controlled text generation aims to generate text with a predefined list of attributes, which could be stylistic or semantic. To reduce the cost of finetuning, recent work of controlled text generation resorts to decoding-time methods, directly making pre-trained models generate texts towards desired attributes during inference. These methods include PPLM~\cite{dathathriplug}, GeDi~\cite{krause2021gedi}, FUDGE~\cite{yang2021fudge}, and DEXPERTS~\cite{liu2021dexperts}.

Controlled text generation is different from personalized generation in that it requires a predefined set of attributes (constraints).

\textit{Text style transfer.}
A task related to controlled text generation is text style transfer. Its goal is to transform a piece of text by controlling certain attributes of the generated text while preserving the content. There are two paradigms: supervised learning using parallel corpora, and unsupervised methods using non-parallel corpora. With parallel data, standard sequence-to-sequence models can be directly employed~\cite{rao2018dear}. There are three approaches when only non-parallel corpora are available. The first approach is to disentangle text into content and attributes for generative modeling~\cite{shen2017style}. The second approach, called prototype editing~\cite{li2018delete}, extracts a sentence template and attribute markers for generation. The third approach constructs pseudo-parallel corpora to train the model~\cite{zhang2018style}.

Unlike text style transfer, personalized generation does not assume that the original text is already given. Like controlled text generation, most methods for text style transfer expect a given set of predefined attributes, which are not available in our setting.

Our work is also aligned with the paradigm of teaching LLMs to reason through chain of thoughts~\cite{wei2022chain, wang2022self}, with our decomposition of tasks deeply inspired by writing education and our model training for each task going beyond prompt engineering.

\section{Problem Formulation}
\label{sec:problem}
We consider the setting where a user is writing a document, which we call the \textbf{current document}. Given the \textbf{immediate context} and the user's \textbf{personal context}, the goal of the personalized model is to complete the document so that the generated document is close to the real current document as if the user finishes writing.

There might be different ways to define the immediate context and the personal context. For simplicity and generality, we use the title and a short start of the current document as the immediate context. The user's personal context is defined as the documents they have written in the past at the time of writing the current document. These contexts are analogous to the sources a student is instructed to write from \cite{cooney2018integrating}. 

Our training task is formulated as follows. Given a list of examples $\{(x_{ut}, \mathcal{D}_{ut}, d_{ut})\}$, where $x_{ut}$ is the immediate context of user $u$ for the current document $d_{ut}$ at time step $t$, and $\mathcal{D}_{ut}=\{d_{u1}, d_{u2}, ..., d_{u,t-1}\}$ is the personal context of past documents, we want to train a personalized generation model $\mathbf{G}$ that generates $d'_{ut}$ based on $(x_{ut}, \mathcal{D}_{ut})$ so that we can maximize the similarity between $d'_{ut}$ and $d_{ut}$. Whenever it is clear from the context, we will omit the subscript $u$ and directly use $x_t$, $\mathcal{D}_{t}$ and $d_t$ instead. We will also use ``user'' and ``author'' interchangeably.

\section{Method Overview}
\label{sec:overview}

\begin{figure}
\centering
\includegraphics[width=0.4\textwidth]{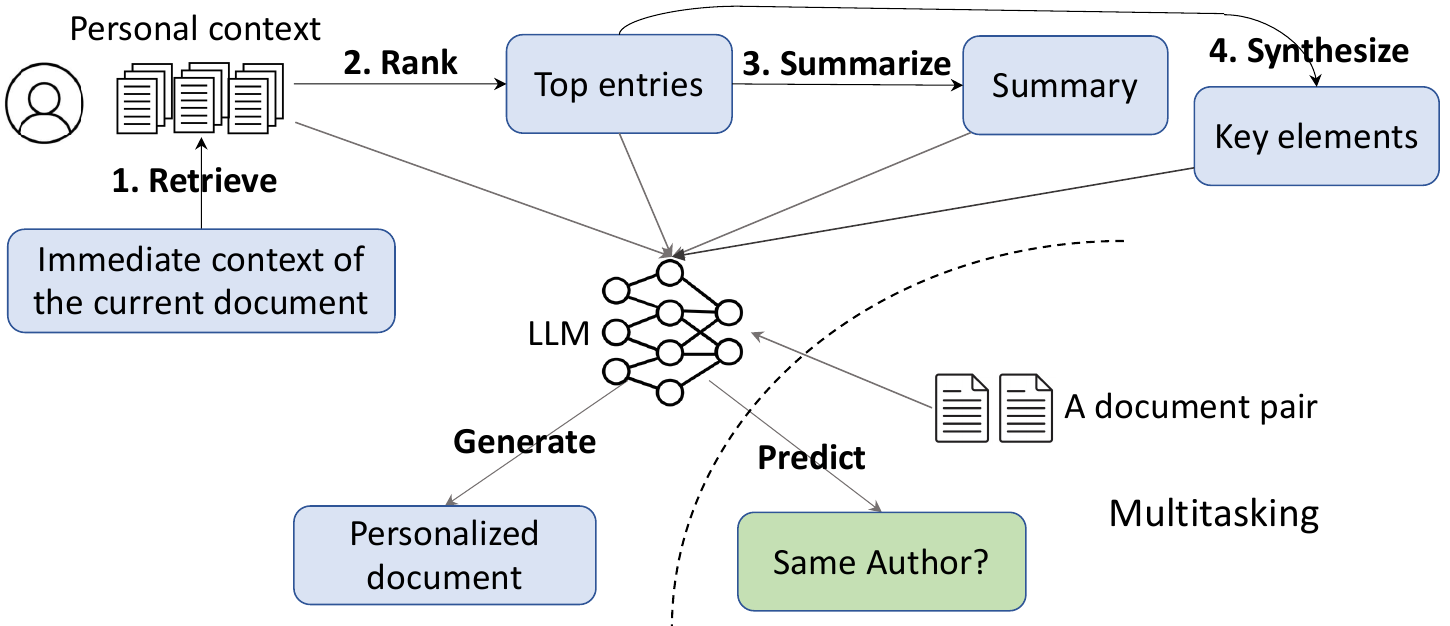}
\caption{The overview of the multistage multitask framework for personalized text generation.}
\label{fig:overview}
\end{figure}

The overview of our multistage multitask framework for personalized text generation is presented in Figure~\ref{fig:overview}.

Given the immediate context $x_t$ of the current document written by a user $u$, a retriever $\mathbf{Re}(x_t, \mathcal{D}_{t})$ retrieves entries from past user document set $\mathcal{D}_{t}$ using $x_t$ as the query. The returned entries are fed to a ranker $\mathbf{Ra}$ to produce ranked entries $\mathcal{E}_{t} = \mathbf{Ra}(\mathbf{Re}(x_t, \mathcal{D}_{t}))$, which are consumed by: (1) a summarization model $\mathbf{Su}(x_t, \mathcal{E}_{t})$ to produce a summary of the multiple documents retrieved; and (2) a synthesis model $\mathbf{Sy}(x_t, \mathcal{E}_{t})$ to produce key elements in these documents. The personalized generation model $\mathbf{G}$ generates the current document $d'_t = \mathbf{G}(x_t, \mathbf{Su}(x_t,\mathcal{E}_{t}), \mathbf{Sy}(x_t,\mathcal{E}_{t}), \mathcal{E}_{t})$ and is trained against the ground-truth current document $d_t$.

We additionally consider an auxiliary task, called author distinction, to help the model better understand the user context and generate better personalized content. Given a document $d_{ui}$ written by a user $u$, we randomly sample another document $d_{vj}$ to form a document pair. The model $\mathbf{G}$ is then trained on a set of tuples $\{(d_{ui}, d_{vj}), y\}$, where the label $y = true$ if $v = u$, otherwise $y = false$. Note that we use text $\{true, false\}$ instead of numerical labels for $y$ since $\mathbf{G}$ is a sequence-to-sequence model.


\section{Personalized text generation}
\label{sec:method}
We discuss the detail of each stage as outlined in Section~\ref{sec:overview}.

\subsection{Retrieval}
In the retrieval stage, given the immediate context $x_t$, the retriever $\mathbf{Re}(x_t, \mathcal{D}_{t})$ uses $x_t$ as the query to retrieve relevant text entries from past document set $\mathcal{D}_{t}$.

To define an immediate context that can be applied to any scenario, we simply use $FirstKCharacters(d_t)$, where \\ $FirstKCharacters(\cdot)$ returns the first $K$ characters of a piece of text. We set $K = 150$ for all experiments. If a document has a title, we concatenate the title and the body as the text.

We experiment with both sparse and dense retrievers to retrieve relevant entries from a user's past documents. We employ BM25~\cite{robertson1995okapi} as the sparse retriever. We use a T5X Retrieval model~\cite{ni2021large, ni2022sentence}, GTR-Large, as our dense retriever. We do not choose models of larger sizes since they demonstrate similar performance but much worse effectiveness-latency trade-offs on benchmark datasets~\cite{ni2021large}.

For dense retrieval, we experiment with two levels of granularity when indexing personal document entries: a \textit{document} level and a \textit{snippet} level. We do not choose a sentence level since many sentences are too short to offer enough context information. We create a snippet in this way: we keep appending sentences from the same document until we reach $250$ characters or we reach the end of the document.

Since the snippets to retrieve are quite short, we only examine the performance of sparse retrieval at the document level.

\subsection{Ranking}
\label{sec:ranking}
Since we experiment with indexing entries at both document and snippet level, we can rank entries accordingly:
\begin{itemize}
    \item \textsc{RankDocBM25}. For sparse retrieval, we retrieve and rank documents based on BM25 scores. 
    \item \textsc{RankDocDense}. For dense retrieval, when we retrieve entries at the document level, we rank retrieved documents based on their embedding similarity with the embedding of the immediate context $x_t$.
    \item \textsc{RankSnippet}. Similarly, for dense retrieval, when we retrieve entries at the snippet level, we rank retrieved snippets based on embedding similarity.
\end{itemize}

During analysis, we find that issues exist for both \textsc{RankDocDense} and \textsc{RankSnippet}. The retrieved results via \textsc{RankDocDense} can be less relevant since embeddings are less effective when the documents to encode are long. 
While for \textsc{RankSnippet}, many similar snippets are retrieved, providing insufficient information for generation. For example, if the immediate context is \textit{I really enjoyed reading the book}, we might retrieve similar snippets like \textit{I enjoy the book}, \textit{The book is fun}, or \textit{I love this book}. They are all relevant but do not provide enough details on why this user enjoys a particular book, which is critical for passage-level generation.

To alleviate the two issues, we propose another dense retrieval strategy, \textsc{RankDocBySnpt}, inspired by past work on using passage evidence in retrieval~\cite{callan1994psg}. \textsc{RankDocBySnpt} retrieves relevant text at the snippet level, which addresses the issue that document embeddings are less effective. At the ranking stage, instead of directly ranking snippets, we rank documents that contain the retrieved snippets, to mitigate lack of diversity in  snippets retrieved via \textsc{RankSnippet}. Specifically for each document $d_i$, we compute the embedding similarity score between each retrieved snippet $s_{ij} \in d_i$ and the immediate context $x_t$, and use the max score as the document score for ranking. That is, $score(d_i, x_t) = \max_{s_{ij} \in d_i}(score(s_{ij}, x_t))$.

To make all the ranking strategies comparable, we concatenate ranked entries into a string and truncate it to $2,500$ characters. Thus the subsequent modules are fed with input text of the same length.

\subsection{Summarization}
\label{sec:sum}
The summarization stage aims to extract important information from the retrieved entries so that the generation model can have a better understanding of what is the most important information in the user's personal context, such as key points, topics, or useful phrases (so they can be reflected in the output). With the summary, the generation model does not need to work on extracting the high-level aspects and generating the exact words at the same time, making the generation task easier.

We experiment with two strategies -- context independent and context dependent summarization. By context dependent, we mean that the summarization is conditioned on the immediate context.

\textit{Context independent summarization}.
We choose a straightforward implementation of context independent summarization -- we finetune an independent LLM, T5-11B~\cite{raffel2020exploring}, on publicly available summarization datasets, and directly use the finetuned model on our ranked entries for inference. The datasets we use are CNN/Daily Mail~\cite{see2017get}, ForumSum~\cite{khalman2021forumsum}, and Reddit TIFU-long~\cite{Kim:2019:NAACL-HLT}.

\begin{figure}
\centering
\includegraphics[width=0.38\textwidth]{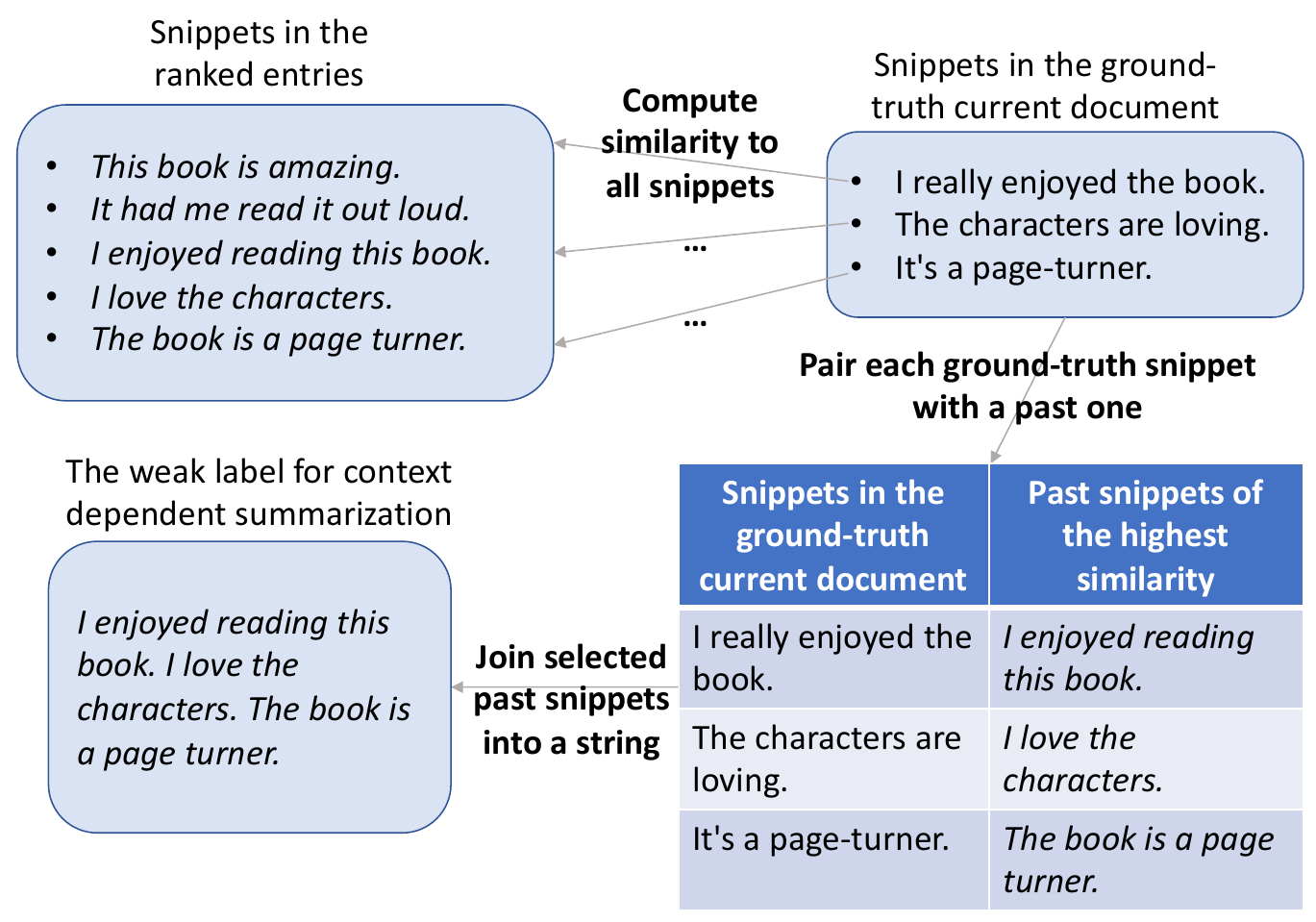}
\caption{Creation of weak labels for context dependent summarization.}
\label{fig:context_sum}
\end{figure}

\textit{Context dependent summarization}.
The challenge to train a context dependent summarization model is the lack of ground-truth labels. We tackle this challenge by generating weak labels based on ground-truth current documents. Our intuition is that we want to extract text from the retrieved results that are more likely to be used in the current document. To this end, we find text from the ranked entries that is similar to the text of the current document, which can be formulated as an extractive summarization task. An example to illustrate this idea can be found in Figure~\ref{fig:context_sum}.


Specifically for each snippet $s_{ti} \in d_t$ in the ground-truth current document, we compute its similarity to all snippets in the ranked entries $\mathcal{E}_{t} = \{e_1, e_2, ..., e_R\}$ retrieved from past documents, where $e_j$ is the $j$-th snippet in the entries $\mathcal{E}_{t}$, and $R$ is the number of snippets we include in the ranking. For simplicity, we reuse the embeddings obtained by the T5X Retrieval model~\cite{ni2021large, ni2022sentence} in the retrieval stage to compute similarity. The snippet with the highest similarity score $e_{max_i} = \arg\max_{e_j \in \mathcal{E}_{t}}(score(e_j, s_{ti}))$ will be added to the candidate snippet list $\mathcal{L}_t$. If the selected snippet $e_{max_i}$ is already in the candidate list $\mathcal{L}_t$, we look for the snippet with the next highest score until we find a new snippet. We keep adding newly selected snippets from the ranked entries to the candidate list until we have iterated through all the snippet pairs.

The weak labels are created by $Join(\mathcal{L}_t)$, which joins the snippets in the candidate list into one string. Given the immediate context $x_t$ and the ranked entries $\mathcal{E}_{t}$, the context dependent summarization model $\mathbf{Su}(x_t, \mathcal{E}_{t})$ is trained using the weak labels $Join(\mathcal{L}_t)$ to minimize the cross-entropy loss. We still choose T5-11B~\cite{raffel2020exploring} as the model. 

One important note is that we only use half of the training data for the generation task to train the summarization model. In this way, the generated summary will be noisier on the unseen half of the training data (similar to the test set). Consequently the generation model will be aware of the noisy summary during training and learn to cope with it by attending to other information. 

Another note is that no validation or test data for the generation task are used to train the summarization model.

\subsection{Synthesis}
\label{sec:synthesis}
The synthesis step aims to find key elements that are common in the top retrieved entries for an overall understanding of the current writing task. We experiment with extracting keywords as the synthesis step in this paper. More sophisticated approaches to synthesis are worth exploration and are left as future work.

Similar to summarization, we investigate two strategies -- context independent and context dependent synthesis. The context also refers to the immediate context.

\textit{Context independent synthesis}.
We extract keywords by finding frequent terms from the past documents $\mathcal{D}_{t}$. We limit terms to unigrams as most users do not have enough documents to extract n-grams with larger n. We also remove stopwords, words with frequency of one, and words with small inverse document frequency (IDF $< 1.5$). We then sort remaining words in descending order by their frequency and then by IDF, and keep up to $20$ words.

\textit{Context dependent synthesis}.
Our idea of creating weak labels for context dependent synthesis is very similar to how we create weak labels for context dependent summarization -- we aim to find important words from the retrieved results that are likely to be used in the generation of the current document. 

To be more specific, for each source word $w_{ti} \in d_t$ in the ground-truth current document, we compute its similarity with each target word $v_{tj} \in \mathcal{E}_{t}$ from the ranked entries. For both source and target words, we skip stopwords or words with IDF $< 1.5$. Two words $(w_{ti}, v_{tj})$ are similar if at least one of the conditions are met:
\begin{itemize}
    \item The two words are identical.
    \item The two words are synonyms as defined in WordNet~\cite{fellbaum1998wordnet}.
    \item The two words are close in the embedding space. We use the uncased GloVe~\cite{pennington2014glove} embeddings pretrained on the Common Crawl dataset. We define two words as similar if their Euclidean distance is less than $4$.
\end{itemize}

We add the qualified target word $v_{tj}$ to the candidate target word list $\mathcal{T}_t$. After going through all possible (source, target) word pairs, the words in the candidate list $\mathcal{T}_t$ are then sorted inversely by the number of times they are selected, and then by IDF. The candidate word list is then joined into a string to form the weak label $Join(\mathcal{T}_t)$.

We still finetune a T5-11B~\cite{raffel2020exploring} model for synthesis. Given the immediate context $x_t$ and the ranked entries $\mathcal{E}_{t}$, the context dependent synthesis model $\mathbf{Sy}(x_t, \mathcal{E}_{t})$ is trained using the weak label $Join(\mathcal{T}_t)$ to minimize the cross-entropy loss.

We use the same set of training examples that are used for summarization. The only difference is that the label is changed from the joined snippet list $Join(\mathcal{L}_t)$ to the joined target word list $Join(\mathcal{T}_t)$.

\subsection{Personalized Generation}

Given the immediate context $x_t$, the ranked entries $\mathcal{E}_{t}$ retrieved from the past documents, the context independent/dependent summary from the summarization model $\mathbf{Su}(x_t, \mathcal{E}_{t})$, the context independent/dependent synthesis from the synthesis model $\mathbf{Sy}(x_t, \mathcal{E}_{t})$, the personalized generation model $\mathbf{G}(x_t, \mathbf{Su}(x_t,\mathcal{E}_{t}), \mathbf{Sy}(x_t,\mathcal{E}_{t}), \mathcal{E}_{t})$ is trained using the ground-truth current document $d_t$ as the label to minimize the cross-entropy loss.

To help the model distinguish between various information sources, we add different prefixes when converting the sources to a single string input. Specifically, the immediate context is prefixed by \textit{passage start}, the summary is prefixed by \textit{summary}, the synthesis is prefixed by \textit{important words}, and the list of ranked entries is prefixed by \textit{past passages}.

\subsection{Multitask Learning}
Studies in language education show that writing and reading skills are highly correlated~\cite{crowhurst1990reading}. Additionally, researches have found that author recognition tasks can be used to assess how much an individual reads~\cite{mccarron2021author}, which correlates with reading proficiency. Inspired by the above studies, in addition to the generation task, which corresponds to writing, we add a reading comprehension task that aims to improve the model's ability to better understand the style of an author. Specifically, we introduce the author distinction task, which requires the model to decide whether a given pair of documents are written by the same author or not.

Given a document $d_{ui}$ written by a user $u$, for half of the time, we randomly sample another document $d_{uj}$ from the same user $u$ to form a positive training example $(x, y) = ((d_{ui}, d_{uj}), true)$. Otherwise, we randomly sample another document $d_{vk}$ from another user $v$ $(v\neq u)$ to form a negative example $(x, y) = ((d_{ui}, d_{vk}), false)$. Note that we use text labels $y \in \{true, false\}$ since the generation model $\mathbf{G}$ is a sequence-to-sequence model.

Since the generation model is simultaneously trained on two tasks, personalized generation and author distinction, we prepend a task-level instruction to the model input to help the model distinguish which task to perform. For the personalized generation task, the instruction is ``\textit{Finish the passage in the user voice}''. While for the author distinction task, the instruction is ``\textit{Predict whether two passages are from the same author}''.

Note that all the documents for the author distinction task are sampled from \textbf{users} that \textbf{never} appear in the validation or test data of the personalized generation task.

\section{Experiment setup}
\label{sec:exp_setup}

In this section, we describe our experiment setup, including datasets, training details, competing methods, and metrics.

\subsection{Datasets}
\label{sec:datasets}
We evaluate our models on three public datasets, each from a representative domain. A summary of data statistics can be found in Table~\ref{tab:data}. For all the three datasets, we give their respective definition of a document.

The Avocado Research Email Collection~\cite{oard2015avocado} consists of emails and attachments taken from $279$ accounts of an information technology company named ``Avocado''. We follow the processing steps of LaMP~\cite{salemi2023lamp} and group emails by sender addresses. Note that the number of senders/users can be more than $279$ accounts as there are senders outside of the company. We treat an email as a document and its subject as the title.

The Amazon review data~\cite{ni2019justifying} provides user reviews from Amazon on different product categories. Since the entire dataset is very large, we choose the biggest category, books, which has the largest number of reviews. We group reviews by users. We treat reviews as documents and review titles as document titles.

The Reddit comments dataset~\cite{reddit2015} consists of all the posts and comments available on Reddit from 2007 to 2015. We treat both posts and comments as documents and group them by users.

For all the three datasets, we deduplicate identical documents from each user's personal context. A document is qualified to be a \textit{current document}, which is the document to be generated, if this document is longer than $300$ characters and the document author has written at least $2$ documents before this one. For each qualified current document, we generate an example. We only keep users who have at least $5$ examples. We include up to $50$ examples per user in case the datasets are dominated by certain active users.

In order to evaluate the model's ability to generalize, we partition the datasets by \textbf{users} so that the validation and test sets only contain documents from users that are unseen in the training set. The partition ratio of users in train/validation/test sets are 85/5/10.

\begin{table*}[!ht]
\centering
\caption{Dataset statistics.}
\label{tab:data}
\begin{tabular}{c|c|c|ccc|ccc}
\toprule
 & \#avg chars of & \#avg past docs &  \multicolumn{3}{c|}{\#users} & \multicolumn{3}{c}{\#current docs (\#examples)}\\
 & current docs & per current doc & Train & Val. & Test & Train & Val. & Test \\
\midrule
Avocado email & 3,648.8 & 42.3 & 501 & 27 & 53 & 13,305 & 764 & 1,227 \\
Amazon review & 1,056.7 & 45.1 & 354,275 & 20,789 & 41,331 & 5,349,661 & 311,469 & 621,438 \\
Reddit & 658.5 & 88.5 & 240,626 & 14,223 & 28,306 & 3,858,731 & 231,307 & 455,582 \\
\bottomrule
\end{tabular}
\end{table*}

\subsection{Training details}
We finetune the T5-11B~\cite{raffel2020exploring} model for all the summarization, synthesis, and personalized generation tasks. The T5-11B models are optimized by the Adafactor algorithm~\cite{shazeer2018adafactor} with a base learning rate of 0.001. We use the first 1,000 training steps for warmup with a linear warmup scheduler. We additionally apply the square root normalized decay of the learning rate. We train the models until the performance converges on the validation set. Beam search~\cite{freitag2017beam} is used for decoding with a beam size of 4.

For multitask learning, we simply mix the examples for personalized generation and author distinction in the ratio of 1:1.

\subsection{Competing methods}
As mentioned in Section~\ref{sec:related}, most personalized generation models are proposed for a particular domain, which relies on domain specific knowledge or features. The work closest to ours is LaMP~\cite{salemi2023lamp}, which employs a retrieval based method for personalization using LLMs. Since LaMP focuses on sentence-length generation, it does not explore the advantage of utilizing snippet- or document-based strategies as we have done in this paper.

We consider the following competing methods for better understanding of credit assignment.

\textit{Baselines}.
We consider these baselines.
\begin{itemize}
    \item \textsc{ImmedCtx}. This method only uses the immediate context $x_t$ as the model input, which is the title and a short start of the current document.
    \item \textsc{UserID}. We add User ID to the model input and train the model using the next snippet generation task. The user ID helps the model memorize the personal style of each user. Since this model performs better on users seen during training, we include all documents that are never used for prediction to train the model, including documents from the validation and the test set. The next snippet generation task requires the model to generate the next snippet given a snippet from a document.
    
    Note that the immediate context includes a short start of the current document, which is also the start of the ground-truth output. Other generation models learn to copy the start to their output to minimize loss but this baseline model is not trained to do so. To make the comparison fair, as postprocess we prepend the start of the current document included in the immediate context to the model output.
    \item \textsc{LLMZeroShot}. We use the PaLM 2 model~\cite{anil2023palm}, which is a new state-of-the-art LLM, for zero-shot generation. PaLM 2 is fed with the input of the \textit{best} configuration based on experiments, including the task instruction to prompt the model to perform the personalized generation task, the immediate context, the context dependent summary, the context dependent synthesis, and the retrieved entries from the best ranking strategy. Similar to the \textsc{UserID} baseline, we also prepend the start of the current document to the model output.
\end{itemize}

\textit{Retrieval augmented methods}.
We experiment with the ranking strategies introduced in Section~\ref{sec:ranking}: one sparse retrieval based method \textsc{RankDocBM25}, and three dense retrieval based methods -- \textsc{RankDocDense}, \textsc{RankSnippet}, and \textsc{RankDocBySnpt}. In addition, we add a recency based baseline, \textsc{RecentDoc}, which retrieves most recent documents written in the past as ranked entries. 

For all these methods, the personalized generation model $\mathbf{G}$ is trained on the immediate context $x_t$ and the ranked entries $\mathcal{E}_{t}$.

\textit{Summarization}.
We examine the summarization methods introduced in Section~\ref{sec:sum} by performing summarization on top of \textit{the best configuration of ranking strategies}. The personalized generation model $\mathbf{G}$ is trained on the immediate context $x_t$, the output from the summarization model $\mathbf{Su}(x_t, \mathcal{E}_{t})$, and the ranked entries $\mathcal{E}_{t}$. We refer to context independent summarization as \textsc{SumCtxInd}, and context dependent summarization as \textsc{SumCtx}.


\textit{Synthesis}.
We evaluate the synthesis methods introduced in Section~\ref{sec:synthesis} by applying synthesis on top of \textit{the best configuration of summarization methods}.  The personalized generation model $\mathbf{G}$ is trained on the immediate context $x_t$, the summary from the summarization model $\mathbf{Su}(x_t, \mathcal{E}_{t})$, the synthesis from the synthesis model $\mathbf{Sy}(x_t, \mathcal{E}_{t})$, and the ranked entries $\mathcal{E}_{t}$. We refer to context independent synthesis as \textsc{SynCtxInd}, and context dependent synthesis as \textsc{SynCtx}.


\textit{Multitask Training}.
We use \textsc{AuthorPred} to refer to the multitask setting, where we add the \textit{author distinction} task on top of the best configuration of the single (generation) task to jointly train the generation model.

\subsection{Metrics}
For each generated current document, we compute its overlap with the ground-truth current document. We adopt \textsc{Bleu}~\cite{papineni2002bleu}, \textsc{Rouge-1}, \textsc{Rouge-2} and \textsc{Rouge-L}~\cite{lin2004rouge} as evaluation metrics, which have been widely used in personalized generation tasks~\cite{li2019towards,salemi2023lamp}.

We conduct statistical significance tests using the paired t-test.

\section{Experimental results}
\label{sec:exp_results}

\subsection{Overall performance}

\begin{table}[h]
\centering
\caption{Overall performance(\%) on the Avocado email dataset. *, \dag \  indicate statistically significant improvement over \textsc{ImmedCtx}, \textsc{RankDocBM25} respectively at the level of 0.01.}
\label{tab:overall_email}
\begin{tabular}{c|c|c|c|c}
\toprule
Avocado email & \textsc{Bleu} & \textsc{Rouge-1} &  \textsc{Rouge-2} & \textsc{Rouge-L} \\
\midrule
\multicolumn{5}{c}{Baselines}\\
\midrule
\textsc{ImmedCtx} & 17.27 & 32.36 & 21.45 & 28.58 \\
\textsc{UserID} & 13.28 & 32.33 & 20.95 & 27.86 \\
\textsc{LLMZeroShot} & 14.93 & $35.06^{*}$ & $22.11^{*}$ & 28.52 \\
\midrule
\multicolumn{5}{c}{Retrieval augmented methods}\\
\midrule
\textsc{RecentDoc} & $19.57^{*}$ & $35.64^{*}$ & $23.96^{*}$ & $31.25^{*}$ \\
\textsc{RankDocBM25} & $21.19^{*}$ & $37.69^{*}$ & $25.99^{*}$ & $33.07^{*}$ \\
\textsc{RankDocDense} & $19.43^{*}$ & $35.62^{*}$ & $23.71^{*}$ & $30.90^{*}$ \\
\textsc{RankSnippet} & $18.69^{*}$ & $35.82^{*}$ & $23.26^{*}$ & $30.78^{*}$ \\
\textsc{RankDocBySnpt} & $21.06^{*}$ & $37.42^{*}$ & $25.65^{*}$ & $32.90^{*}$ \\
\midrule
\multicolumn{5}{c}{+Summarization}\\
\midrule
\textsc{SumCtxInd} & $21.23^{*}$ & $37.58^{*}$ & $25.79^{*}$ & $33.15^{*}$ \\
\textsc{SumCtx} & $23.17^{*\dag}$ & $39.31^{*\dag}$ & $26.64^{*\dag}$ & $34.37^{*\dag}$ \\
\midrule
\multicolumn{5}{c}{+Synthesis}\\
\midrule
\textsc{SynCtxInd} & $23.06^{*\dag}$ & $39.24^{*\dag}$ & $26.72^{*\dag}$ & $34.52^{*\dag}$ \\
\textsc{SynCtx} & \boldsymbol{$23.44^{*\dag}$} & $40.38^{*\dag}$ & $26.93^{*\dag}$ & $34.34^{*\dag}$ \\
\midrule
\multicolumn{5}{c}{+Multitask}\\
\midrule
\textsc{AuthorPred} & $23.27^{*\dag}$ & \boldsymbol{$41.02^{*\dag}$} & \boldsymbol{$28.60^{*\dag}$} & \boldsymbol{$35.70^{*\dag}$} \\
\bottomrule
\end{tabular}
\end{table}

\begin{table}[!ht]
\centering
\caption{Overall performance(\%) on the Amazon review dataset. *, \dag \  indicate statistically significant improvement over \textsc{ImmedCtx}, \textsc{RankDocBySnpt} respectively at the level of 0.01.}
\label{tab:overall_amazon}
\begin{tabular}{c|c|c|c|c}
\toprule
Amazon review & \textsc{Bleu} & \textsc{Rouge-1} &  \textsc{Rouge-2} & \textsc{Rouge-L} \\
\midrule
\multicolumn{5}{c}{Baselines}\\
\midrule
\textsc{ImmedCtx} & 17.83 & 36.22 & 22.49 & 31.77 \\
\textsc{UserID} & 17.61 & 36.62 & 22.85 & $32.11^{*}$ \\
\textsc{LLMZeroShot} & 16.29 & $38.74^{*}$ & 22.52 & 30.79 \\
\midrule
\multicolumn{5}{c}{Retrieval augmented methods}\\
\midrule
\textsc{RecentDoc} & $19.09^{*}$ & $37.51^{*}$ & $23.37^{*}$ & $32.67^{*}$ \\
\textsc{RankDocBM25} & $19.49^{*}$ & $37.79^{*}$ & $23.56^{*}$ & $32.85^{*}$ \\
\textsc{RankDocDense} & $19.38^{*}$ & $37.49^{*}$ & $22.92^{*}$ & $32.48^{*}$ \\
\textsc{RankSnippet} & $19.44^{*}$ & $37.45^{*}$ & $23.10^{*}$ & $32.50^{*}$ \\
\textsc{RankDocBySnpt} & $19.35^{*}$ & $38.28^{*}$ & $23.87^{*}$ & $33.23^{*}$ \\
\midrule
\multicolumn{5}{c}{+Summarization}\\
\midrule
\textsc{SumCtxInd} & $19.17^{*}$ & $38.33^{*}$ & $23.66^{*}$ & $33.35^{*}$ \\
\textsc{SumCtx} & $19.81^{*\dag}$ & $38.66^{*}$ & $24.25^{*\dag}$ & $33.57^{*}$ \\
\midrule
\multicolumn{5}{c}{+Synthesis}\\
\midrule
\textsc{SynCtxInd} & $19.84^{*\dag}$ & $38.68^{*}$ & $24.31^{*\dag}$ & $33.61^{*}$ \\
\textsc{SynCtx} & \boldsymbol{$19.87^{*\dag}$} & \boldsymbol{$39.46^{*\dag}$} & \boldsymbol{$24.66^{*\dag}$} & \boldsymbol{$33.97^{*\dag}$} \\
\midrule
\multicolumn{5}{c}{+Multitask}\\
\midrule
\textsc{AuthorPred} & $19.78^{*\dag}$ & $39.36^{*\dag}$ & $24.56^{*\dag}$ & $33.87^{*\dag}$ \\
\bottomrule
\end{tabular}
\end{table}

\begin{table}[!hb]
\centering
\caption{Overall performance(\%) on the Reddit dataset. *, \dag \ indicate statistically significant improvement over \textsc{UserID}, \textsc{RankSnippet} respectively at the level of 0.01.}
\label{tab:overall_reddit}
\begin{tabular}{c|c|c|c|c}
\toprule
Reddit & \textsc{Bleu} & \textsc{Rouge-1} &  \textsc{Rouge-2} & \textsc{Rouge-L} \\
\midrule
\multicolumn{5}{c}{Baselines}\\
\midrule
\textsc{ImmedCtx} & 26.33 & 43.69 & 33.52 & 40.53 \\
\textsc{UserID} & 26.55 & $44.71$ & 34.61 & $41.53$ \\
\textsc{LLMZeroShot} & 24.04 & 43.67 & 31.24 & 37.85 \\
\midrule
\multicolumn{5}{c}{Retrieval augmented methods}\\
\midrule
\textsc{RecentDoc} & $28.23^{*}$ & $44.83$ & $34.57$ & $41.72$ \\
\textsc{RankDocBM25} & $28.97^{*}$ & $45.71^{*}$ & $35.13^{*}$ & $42.52^{*}$ \\
\textsc{RankDocDense} & $28.82^{*}$ & $45.61^{*}$ & $35.22^{*}$ & $42.56^{*}$ \\
\textsc{RankSnippet} & $29.08^{*}$ & $45.94^{*}$ & $35.39^{*}$ & $42.68^{*}$ \\
\textsc{RankDocBySnpt} & $28.87^{*}$ & $45.67^{*}$ & $35.24^{*}$ & $42.54^{*}$ \\
\midrule
\multicolumn{5}{c}{+Summarization}\\
\midrule
\textsc{SumCtxInd} & $28.91^{*}$ & $45.71^{*}$ & $35.26^{*}$ & $42.57^{*}$ \\
\textsc{SumCtx} & $28.92^{*}$ & $46.09^{*}$ & $35.82^{*\dag}$ & $43.14^{*\dag}$ \\
\midrule
\multicolumn{5}{c}{+Synthesis}\\
\midrule
\textsc{SynCtxInd} & $28.82^{*}$ & $45.91^{*}$ & $35.87^{*\dag}$ & $43.09^{*\dag}$ \\
\textsc{SynCtx} & \boldsymbol{$29.19^{*}$} & $46.45^{*\dag}$ & $36.13^{*\dag}$ & $43.27^{*\dag}$ \\
\midrule
\multicolumn{5}{c}{+Multitask}\\
\midrule
\textsc{AuthorPred} & $29.13^{*}$ & \boldsymbol{$47.11^{*\dag}$} & \boldsymbol{$36.94^{*\dag}$} & \boldsymbol{$43.95^{*\dag}$} \\
\bottomrule
\end{tabular}
\end{table}

The overall performance on the three datasets are listed in Table~\ref{tab:overall_email}, \ref{tab:overall_amazon}, and \ref{tab:overall_reddit} respectively. In general, retrieval augmented methods perform better than baselines that do not utilize the retrieved information. Summarization and synthesis bring additional gains when they are dependent on the immediate context. Multitask learning further improves the model's generation ability in most datasets. We compare important methods in detail below.

\textit{Comparison among baselines}.
Comparing with \textsc{ImmedCtx}, \textsc{UserID} performs surprisingly well, especially on the Amazon review and the Reddit data. This is because the \textsc{UserID} model is trained on the next snippet prediction task, resulting in relatively shorter generated output. By memorizing the user IDs, the model has an advantage on datasets with shorter document length, e.g., the Amazon review and the Reddit data. However, since this model requires user IDs as input, it has problems in generalizing to new users or scaling to a large number of users.

\textsc{LLMZeroShot} is provided with the same input as \textsc{SynCtx}. The reduced performance suggests that finetuning is critical to the personalized generation task.

\textit{Retrieval augmented methods}.
All the retrieval augmented methods outperform \textsc{ImmedCtx}, which excludes all documents from a user's personal context. This indicates that past documents provide useful information when the model generates the current document for a user.

\textsc{RecentDoc} unexpectedly performs on par with some similarity based retrieval methods in many cases, especially on the Avocado email dataset. But it still performs worse than \textsc{RankDocBySnpt}. On one hand, \textsc{RecentDoc} is a decent strategy as a user's recent documents might share similar writing style, and even similar content, e.g., a user is writing multiple emails concerning a particular topic, or a user starts to take interest in a particular book genre and are thus writing similar book reviews. On the other hand, providing information more relevant to the current topic based on the immediate context using \textsc{RankDocBySnpt} instead of simply retrieving the most recent content can further improve the generation performance. 

\textsc{RankDocBM25} performs closely to \textsc{RankDocBySnpt} except for the Amazon review dataset, where \textsc{RankDocBM25} is less effective. This suggests that BM25 is still a powerful retrieval strategy even when the query, which is the immediate context, is longer than common search queries and is written in natural language.

\textsc{RankDocBySnpt} consistently performs better than or equally well as other retrieval based methods by combining the strength of \textsc{RankDocDense} and \textsc{RankSnippet}. By retrieving on the snippet level, dense retrieval becomes more effective by encoding less information in a single embedding. By ranking documents that contain the retrieved snippets, the generation model is able to extract more diverse or detailed information for content generation. \textsc{RankDocDense}, \textsc{RankSnippet} and \textsc{RankDocBySnpt} all perform well on the Reddit data, due to the relatively short document length of this dataset.

\textit{Summarization}.
\textsc{SumCtxInd} performs similarly as retrieval augmented methods without the summarization step, while \textsc{SumCtx} outperforms retrieval augmented methods. This indicates that a generic summary does not provide additional information to the generation model. The summary is more useful when it considers the immediate context. It guides the generation model to the incorporation of possible topics or phrases when generating the current document. Without the context dependent summary, the generation model needs to consider the high-level topics and the exact words to generate at the same time, making the task harder.

\textit{Synthesis}.
We observe patterns similar as summarization -- \textsc{SynCtxInd} performs on par with \textsc{SumCtx}, a method without synthesis, while \textsc{SynCtx} outperforms \textsc{SumCtx}. This means that synthesis is useful only when it is dependent on the immediate context. We also see that \textsc{SynCtx} improves the \textsc{Rouge-1} metric more than other metrics. This is due to our design choice of the synthesis stage by identifying important unigrams. More sophisticated synthesis methods are worth exploring as future work to improve metrics other than \textsc{Rouge-1}.

\textit{Multitask}.
By jointly learning to predict whether two documents are from the same author, \textsc{AuthorPred} performs better than \textsc{SynCtx}, which has a single task setting, on the Avocado email and the Reddit datasets, and performs closely on the Amazon review dataset. This indicates that improving the model's reading ability by differentiating the writing style of different authors does no harm to the generation performance, which often benefits.

\subsection{Formulation of input and output}
Since Table~\ref{tab:overall_email}, \ref{tab:overall_amazon}, and \ref{tab:overall_reddit} have already included the ablation study by reporting the performance of adding one new component at a time, we additionally investigate whether varying the formulation of the input and output will lead to significant change in performance. We study the following settings.
\begin{itemize}
    \item \textsc{NoRankedEntries}. We investigate whether the ranked entries are still necessary when the summary and the synthesized outcome are available. To this end, we train the generation model using the immediate context, the context dependent summary, and the context dependent synthesis.
    \item \textsc{RemoveDocStart}. The immediate context includes a short start of the current document. For all the previous experiments, this short start is still present in the ground-truth label, which is the current document, when we train the generation model. We study the effect of removing the short start from the ground-truth label. Note that to make the comparison fair, we prepend the short start to the model generated text when computing metrics.
    \item \textsc{ImmedCtxAtEnd}. In all the experiments above, the information sources are placed in the order of the immediate context, the summary, the synthesis, and the ranked entries. We study whether the order matters by placing the immediate context at the end.
\end{itemize}

\begin{table}[h]
\centering
\caption{Analysis of formulation of input and output on the Avocado email dataset based on absolute change of performance(\%) over \textsc{AuthorPred}. Other datasets show similar patterns and are omitted to save space.}
\label{tab:input_output}
\begin{tabular}{c|c|c|c|c}
\toprule
Avocado email & \textsc{Bleu} & \textsc{Rouge-1} &  \textsc{Rouge-2} & \textsc{Rouge-L} \\
\midrule
\textsc{NoRankedEntries} & -2.53 & -2.6 & -3 & -2.75 \\
\textsc{RemoveDocStart} & -0.85 & -0.74 & -2.25 & -1.78 \\
\textsc{ImmedCtxAtEnd} & -0.12 & 0.05 & 0.13 & -0.08 \\
\bottomrule
\end{tabular}
\end{table}

\begin{table}[h]
\small
\centering
\caption{Examples of model output. Italic words are used in the immediate context.}
\label{tab:case_studies}
\begin{tabular}{p{0.47\textwidth}}
\toprule
\textbf{Ground-truth doc}: \textit{I just did not enjoy this book. I discovered this author a few weeks ago and have been making my way through her} cocktail series but this one was not good for me. Didnt particularly like the characters (and I get it, Lucas looks like Prince Harry) and the plot was slow and less than engaging. I finished it, but only because Im stubborn. \\
\hline
\textbf{\textsc{SumCtx} output}: \textit{I just did not enjoy this book. I discovered this author a few weeks ago and have been making my way through her} books. I really liked the first book in the series, but this one was just not my favorite. I just didn't find myself connecting with the characters. I found the heroine to be a bit too naive and I just didn't find myself falling for the hero. \\
\hline
\textbf{\textsc{RankDocBySnpt} output}: \textit{I just did not enjoy this book. I discovered this author a few weeks ago and have been making my way through her} books. I've enjoyed the first two books in this series and I'm not sure why I didn't like this one. I'm not sure if it's because I'm not a fan of the whole "friends to lovers" thing, but I just didn't like the way this book was written. \\
\hline
\textbf{Why \textsc{SumCtx} is better}: the summarization model mentions \textit{I didn't fall for the hero}, which makes the generation model comment more on the characters.\\
\hline\hline
\textbf{Ground-truth doc}: \textit{Thanks for your insight. I do websites, mobile applications, desktop applications. I have a lot of success with odesk and have been using it ever} since. I have been thinking of developing stuff for my own the way I do it for my clients. I just get too busy with side projects and my full-time employment. \\
\hline
\textbf{\textsc{SynCtx} output}: \textit{Thanks for your insight. I do websites, mobile applications, desktop applications. I have a lot of success with odesk and have been using it ever} since. I have a lot of clients and I have been using it for a long time. I am a professional developer though and I have been doing this for a long time. \\
\hline
\textbf{\textsc{SumCtx} output}: \textit{Thanks for your insight. I do websites, mobile applications, desktop applications. I have a lot of success with odesk and have been using it ever} since. I have a lot of projects. I have a lot of experience with php, javascript, html, css, javascript, html5, c\#, javascript, c\# \\
\hline
\textbf{Why \textsc{SynCtx} is better}: the synthesis model mentions \textit{developer} and \textit{freelancer} so that the generation model can focus on the topic of development and clients.\\
\hline\hline
\textbf{Ground-truth doc}: \textit{As another reviewer wrote, this book flows gently. There isn't a lot of action and yes, the timeline bounces around but it's} not hard to figure out what is happening. But fair warning - this is a sad story. I got choked up once or twice. It's very well written but the story isn't really groundbreaking. \\
\hline
\textbf{\textsc{AuthorPred} output}: \textit{As another reviewer wrote, this book flows gently. There isn't a lot of action and yes, the timeline bounces around but it's} not confusing. The characters are well-developed and the story is interesting. I did find the ending a little sad, but I guess that's what I was expecting. \\
\hline
\textbf{\textsc{SynCtx} output}: \textit{As another reviewer wrote, this book flows gently. There isn't a lot of action and yes, the timeline bounces around but it's} not confusing. The story is about a group of friends who have been friends since childhood. They are all in their late 30s and have been friends since childhood. \\
\hline
\textbf{Why \textsc{AuthorPred} is better}: The author distinction task helps the model understand that this user often gives a high-level review of the story instead of going to details.\\
\bottomrule
\end{tabular}
\end{table}

Due to the space limit, we only show the performance on the Avocado email data in Table~\ref{tab:input_output}. Other datasets show similar patterns.

\textsc{NoRankedEntries} underperforms \textsc{AuthorPred} by a large margin, meaning that the generation model still relies on the retrieved entries for information, e.g., word usage and writing style, even when the summary and the synthesis are available. There is a performance degradation of \textsc{RemoveDocStart}. We suspect that the presence of the short start of the current document in both the input and the ground-truth label helps the model understand the task better. This is supported by the observation that the model converges faster during training when the short start is included in the ground-truth label. The close performance between \textsc{ImmedCtxAtEnd} and \textsc{AuthorPred} indicates that reordering the information sources does not affect the performance at least in the case of finetuning.

\subsection{Case studies}

We provide some illustrative examples in Table~\ref{tab:case_studies} for a better understanding of why the proposed method could provide better guidance on how to generate personalized content.

\section{Conclusion}
We propose a general approach for teaching large language models for personalized text generation. Analogous to how students are instructed to write from sources in a sequence of steps, the proposed approach consists of multiple stages: retrieval, ranking, summarization, synthesis, and generation. Additionally, inspired by the observation that reading and writing skills are correlated, we create a multitask setting that improves the model's reading ability by distinguishing the authorship of given document pairs. This multitask setting further improves the model's ability to generate personalized text empirically. We evaluate our models on three publicly released datasets from representative domains. Our results demonstrate the effectiveness of the multistage multitask framework. Investigation into the incorporation of world knowledge, e.g., product information, is a promising direction for future work.



\newpage

\balance
\bibliographystyle{ACM-Reference-Format}
\bibliography{bib}



\end{document}